\documentclass[runningheads]{llncs}
\usepackage{graphicx}
\usepackage{amsmath}
\usepackage[misc]{ifsym}
\usepackage{multicol}
\usepackage{multirow}
\usepackage{threeparttable}
\usepackage{todonotes}

\begin{document}

\title{A Self-Supervised Deep Framework for Reference Bony Shape Estimation in Orthognathic Surgical Planning}
\titlerunning{Self-Supervised Orthognathic Reference Shape Estimation}

\author{Deqiang Xiao\inst{1} \and	
Hannah Deng\inst{2} \and	
Tianshu Kuang\inst{2} \and	
Lei Ma\inst{1} \and	
Qin Liu\inst{1} \and	 
Xu Chen\inst{1} \and	
Chunfeng Lian\inst{1} \and
Yankun Lang\inst{1} \and	
Daeseung Kim\inst{2} \and	
Jaime Gateno\inst{2,3} \and	
Steve Guofang Shen\inst{4} \and	
Dinggang Shen\inst{1} \and	
Pew-Thian Yap\inst{1}\textsuperscript{,\,\Letter} \and	
James J. Xia\inst{2,3}\textsuperscript{,\,\Letter}	
}	
% index{Xiao, Deqiang}
% index{Deng, Hannah}
% index{Kuang, Tianshu}
% index{Ma, Lei}
% index{Liu, Qin}
% index{Chen, Xu}
% index{Lian, Chunfeng}
% index{Lang, Yankun}
% index{Kim, Daeseung}
% index{Gateno, Jaime}
% index{Shen, Steve Guofang}
% index{Shen, Dinggang}
% index{Yap, Pew-Thian}
% index{Xia, James J.}

\authorrunning{D. Xiao et al.}

\institute{Department of Radiology and Biomedical Research Imaging Center (BRIC), University of North Carolina at Chapel Hill, NC 27599, USA\\
 \and
Department of Oral and Maxillofacial Surgery, Houston Methodist Hospital, TX 77030, USA \\
 \and
Department of Surgery (Oral and Maxillofacial Surgery), Weill Medical College, Cornell University, NY 10065, USA \and
Oral and Craniomaxillofacial Surgery at Shanghai Ninth Hospital, Shanghai Jiaotong University College of Medicine, Shanghai 200011, China
\email{ptyap@med.unc.edu}, \email{jxia@houstonmethodist.org}
}
\maketitle

\begin{abstract}
Virtual orthognathic surgical planning involves simulating surgical corrections of jaw deformities on 3D facial bony shape models. Due to the lack of necessary guidance, the planning procedure is highly experience-dependent and the planning results are often suboptimal. A reference facial bony shape model representing normal anatomies can provide an objective guidance to improve planning accuracy. Therefore, we propose a self-supervised deep framework to automatically estimate reference facial bony shape models. Our framework is an end-to-end trainable network, consisting of a simulator and a corrector. In the training stage, the simulator maps jaw deformities of a patient bone to a normal bone to generate a simulated deformed bone. The corrector then restores the simulated deformed bone back to normal. In the inference stage, the trained corrector is applied to generate a patient-specific normal-looking reference bone from a real deformed bone. The proposed framework was evaluated using a clinical dataset and compared with a state-of-the-art method that is based on a supervised point-cloud network. Experimental results show that the estimated shape models given by our approach are clinically acceptable and significantly more accurate than that of the competing method.

\end{abstract}

\section{Introduction}
\label{sec:introduction}
\begin{figure}[!t]
	\centerline{\includegraphics[width=0.9\textwidth]{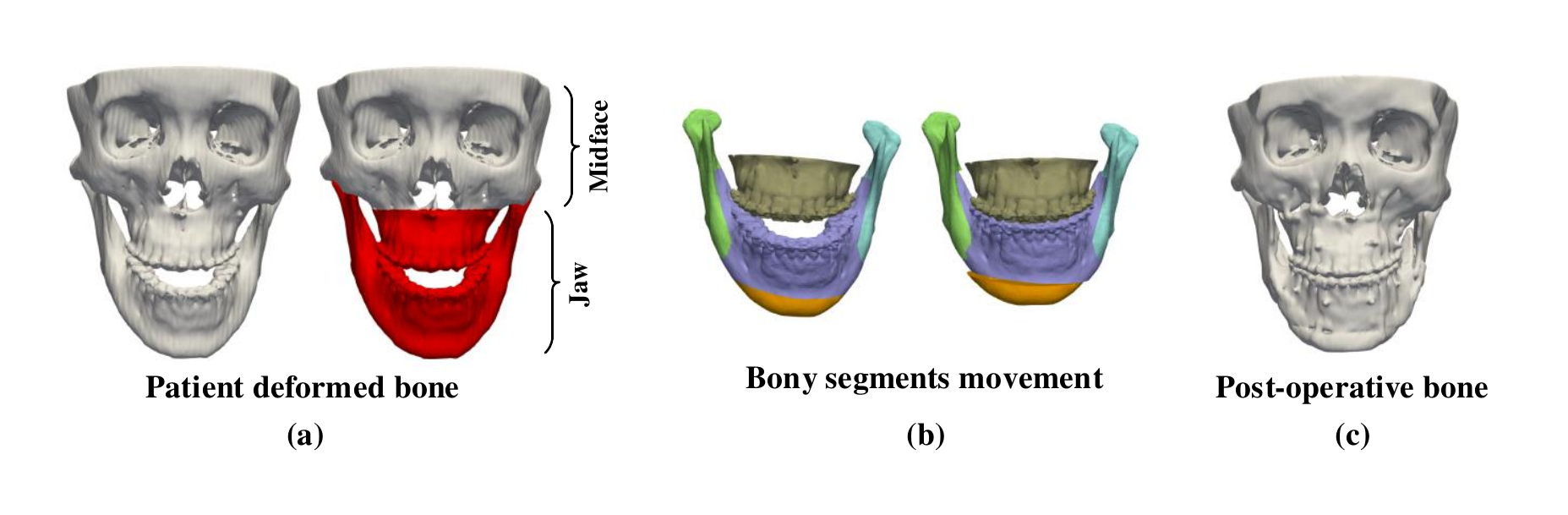}}
	\caption{(a) The bony surface of a patient with jaw deformity, consisting of a normal ``midface'' and a deformed ``jaw'' (in red). (b) The jaw is cut into several segment pieces and moved to form a new normal-looking shape model. (c) The patient's facial bony shape model after treatments.} 
	\label{fig:surgical_planning}
\end{figure}

Orthognathic surgery requires a detailed surgical plan to correct jaw deformities. During computer-aided orthognathic surgical planning, a patient’s head computed tomography (CT) or cone-beam computed tomography (CBCT) scan is acquired to generate three-dimensional (3D) craniomaxillofacial (CMF) bony shape models~\cite{Xia2015a}. The deformed maxilla and mandible (i.e., ``jaw'' in Fig.~\ref{fig:surgical_planning}(a)) are then osteotomized from the 3D models into multiple bony segments (Fig.~\ref{fig:surgical_planning}(b)). Under the guidance of 3D cephalometric analysis~\cite{Xia2015b}, a surgeon discretely moves each bony segment to a desired position, thus forming a new normal-looking CMF skeletal anatomy. A group of surgical splints are then designed and fabricated to correct the patient's jaw according to the plan during the surgery (Fig.~\ref{fig:surgical_planning}(c)). Cephalometric analysis is based on a group of specific distances and angles calculated from the anatomical landmarks on the 3D models, these measurements provide a limited guidance to the planning process, forming an ``ideal'' surgical plan that still heavily relies on the surgeon's clinical experience and imagination of what the patient’s normal bone should look like. Therefore, from the surgeon’s perspective, a reference CMF bony shape model is highly desirable to accurately guide the movement of the bony segments. 

Wang et al.~\cite{Wang2015} proposed to estimate patient-specific reference bony shapes based on sparse representation. Specifically, they represented a patient’s bony shape with a set of landmarks, and split them into midface landmarks and jaw landmarks. By representing the patient’s midface landmarks with a midface dictionary from normal subjects, a set of sparse coefficients were obtained, and applied to a normal jaw dictionary to estimate the patient’s normal jaw landmarks. The reference bony shape model was then generated based on the estimated landmarks. This method can work well in certain circumstances, but is heavily dependent on a small group of pre-digitized landmarks, causing the estimation performance to be sensitive to the locations of landmarks. To alleviate this problem, a recent method~\cite{Xiao2021} was introduced to first simulate paired deformed-normal bony surfaces based on sparse representation, and then train a supervised point-cloud network using the simulated data to generate patient-specific reference bony surfaces. Although this network~\cite{Xiao2021} outperforms the method in~\cite{Wang2015}, its generalizability is limited by the extent deformity can be simulated. 

We hypothesize that a fully end-to-end trainable network is capable of estimating reference CMF bony shapes, and the network can be directly trained using random pairs of deformed-normal bones. We propose a self-supervised deep framework to first map jaw deformities of a patient bone to a normal bone to generate a simulated deformed bone with a simulator network, and then restore the simulated deformed bone back to normal with a corrector network. We train the framework using a series of bony surfaces from patients and normal subjects. The trained corrector is applied to patients to generate patient-specific reference shape models. The proposed framework was evaluated using a clinical dataset and showed superior performance over the method that is based on supervised learning \cite{Xiao2021}. 

\section{Methods}
\label{sec:method}
\begin{figure}[!t]
	\centerline{\includegraphics[width=0.9\textwidth]{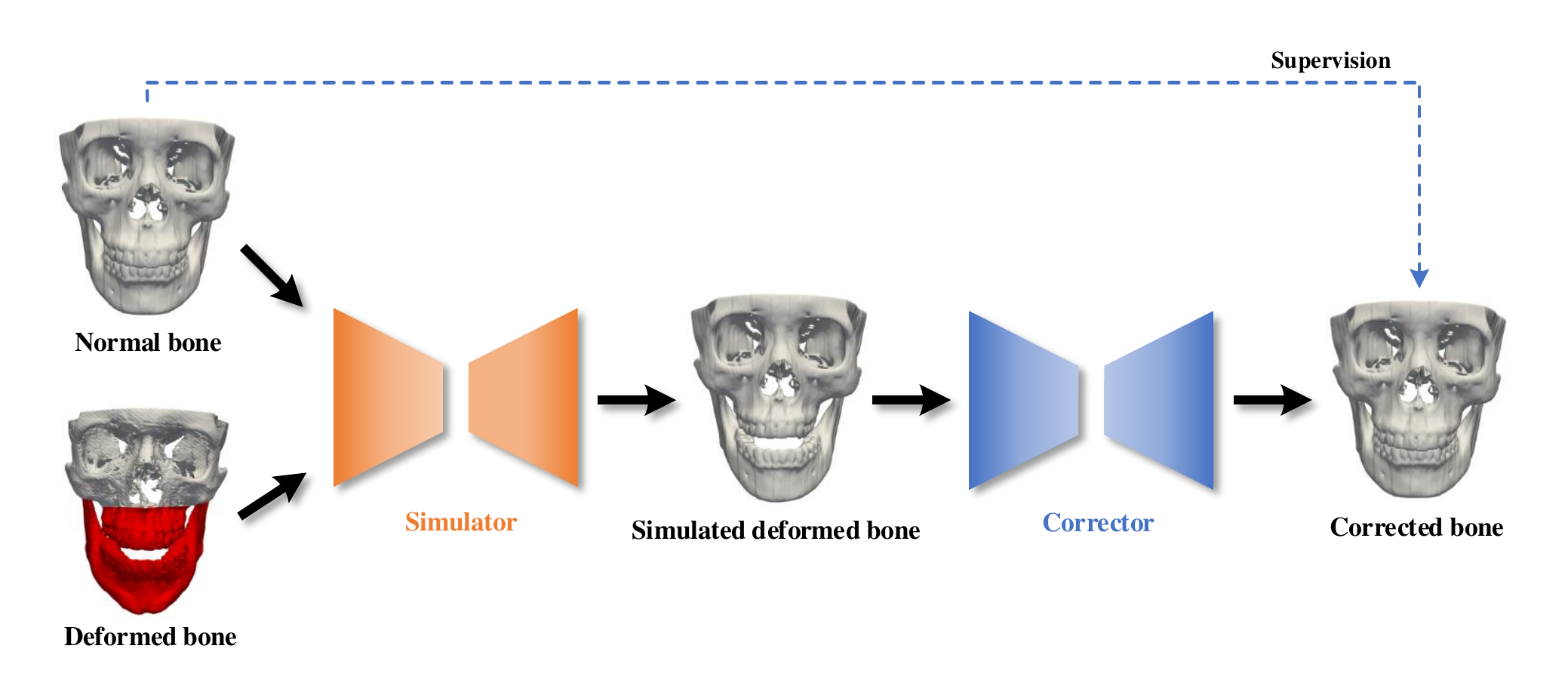}}
	\caption{An overview of our self-supervised deep framework for estimating reference CMF bony shape models.} 
	\label{fig:method_outline}
\end{figure}

An overview of our framework is provided in Fig.~\ref{fig:method_outline}. Given any pair of deformed-normal bony surfaces, the simulator network takes as input the coordinates of $N$ vertices from the two surfaces and outputs the vectors that shift the vertices of the normal bony surface to generate the simulated deformed bony surface with the jaw similar to the input deformed bone and the midface same as the input normal bone. 
The corrector network learns a set of displacement vectors from the vertex coordinates of the simulated deformed bony surface and generates the corrected bony surface that matches the input normal bony surface.

\subsubsection{Simulator Network:}
\label{sec:simulator_network}
\begin{figure}[!t]
	\centerline{\includegraphics[width=1.0\textwidth]{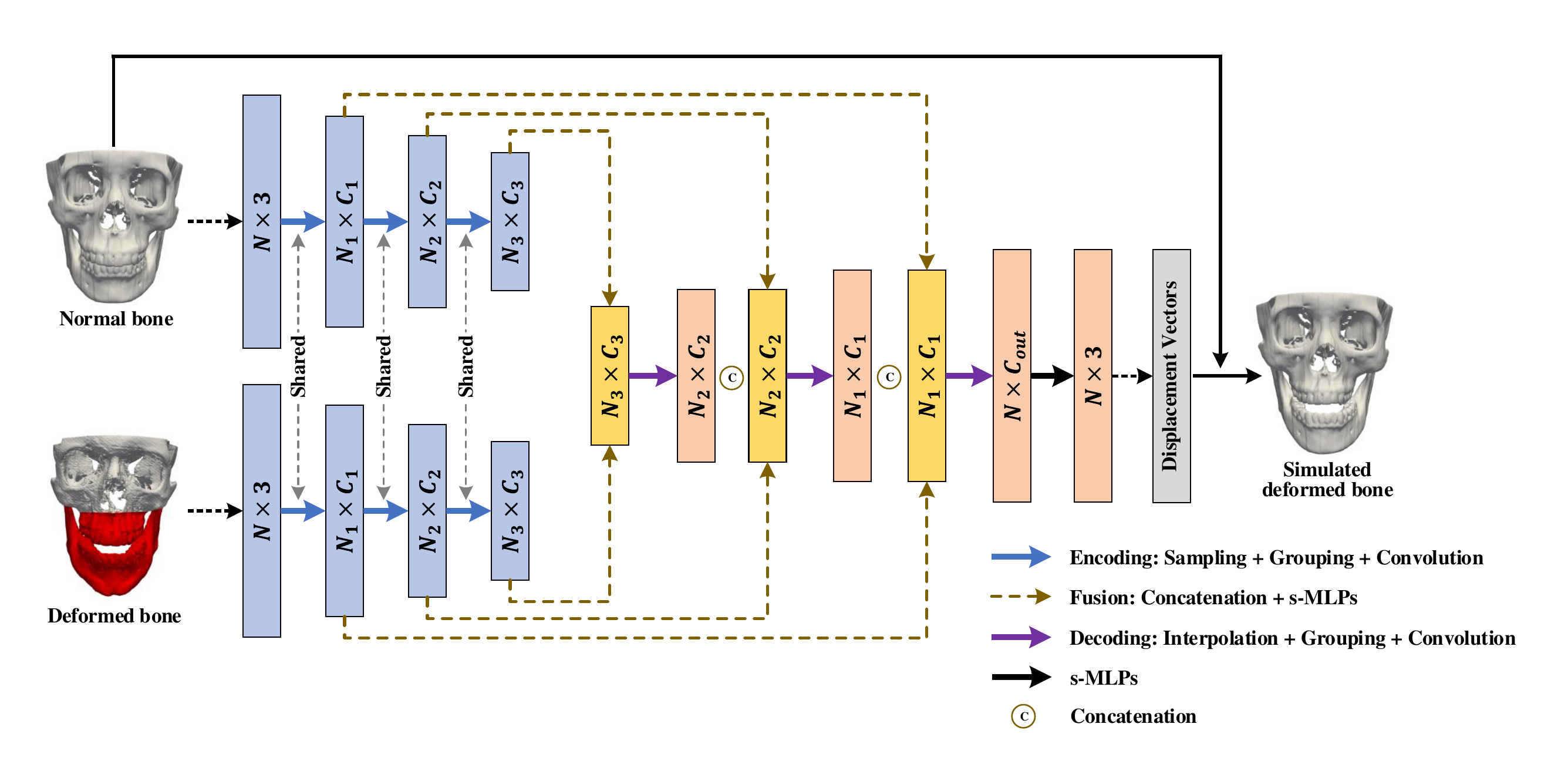}}
	\caption{The architecture of the simulator network, consisting of point-feature encoding, fusion, and decoding layers.} 
	\label{fig:simulator_network}
\end{figure}

The simulator network consists of a series of encoding, fusion, and decoding layers to learn point features from the coordinates of vertices (Fig.~\ref{fig:simulator_network}). There are two encoding branches with shared weights to extract point features from the two input bony surfaces separately. Each encoding layer selects a subset of $N_{\text{sub}}$ points from the input points via furthest point sampling~\cite{Qi2017a}. For each sampled point, its neighboring points in a 3D ball region with radius $r$ are gathered from the input points. The point convolution is then performed over point features of the gathered points via PointConv~\cite{Wu2019}, which is implemented based on shared multi-layer perceptron (s-MLP) and max-pooling~\cite{Qi2017b}. Following the encoding layers, the outputs of corresponding layers from two encoding branches are fused via concatenation and s-MLPs. The fused features are then decoded by upsampling via point interpolation~\cite{Qi2017a}, grouping via 3D ball neighboring, and convolution via PointConv. The decoding layer increases the number $N_{\text{sub}}$ of points and reduces the dimension $C_{\text{feature}}$ of each feature vector. A set of cascaded fusion and decoding layers are applied to generate a $N \times 3$ output vector $V_{\text{simulator}}$, which updates the positions of the vertices of the input normal bone to obtain the simulated deformed bone. 

We train the simulator network using loss function
\begin{equation} \label{eq:1}
L_{\text{simulator}}=L_{\text{jaw}}+L_{\text{midface}}+\alpha L_{\text{smooth}}+\beta L_{\text{simu\_reg}},
\end{equation}
which encourages the jaw of the simulated deformed bone to be similar to the input deformed bone based on the relative coordinates of vertices. Specially, a set of landmarks are defined on each bony surface, the relative coordinate vector $r_{\text{def\_jaw}}(i,j)$ of the $i$-th jaw vertex relative to the $j$-th landmark on the deformed bony surface is calculated as
\begin{equation} \label{eq:2}
r_{\text{def\_jaw}}(i,j)=c_{\text{def\_jaw}}(i)-c_{\text{def\_landmark}}(j),
\end{equation} 
where $c_{\text{def\_jaw}}(i)$ and $c_{\text{def\_landmark}}(j)$ are respectively the coordinate vectors of the $i$-th vertex and the $j$-th landmark. Similarly, the relative coordinate vector for the $i$-th jaw vertex of the simulated deformed bony surface is
\begin{equation} \label{eq:3}
r_{\text{simu\_jaw}}(i,j)=c_{\text{simu\_jaw}}(i)-c_{\text{simu\_landmark}}(j).
\end{equation} 
We calculate $L_{\text{jaw}}$ by
\begin{equation} \label{eq:4}
L_{\text{jaw}}=\frac{1}{N_{\text{jaw}}}\sum_{i=1}^{N_{\text{jaw}}}\sum_{j=1}^{K}\left \| r_{\text{def\_jaw}}(i,j)-r_{\text{simu\_jaw}}(i,j) \right \|_{2},
\end{equation} 
where $N_{\text{jaw}}$ is the number of vertices that belong to the jaw, $K$ is the number of landmarks, and $\left \| \cdot  \right \|_{2}$ is the  $\ell_2$-norm. 
$L_{\text{midface}}$ forces the network to fix the midface during simulation. To keep the midface unchanged, i.e., $L_{\text{midface}}=0$, we set the displacement vectors corresponding to midface vertices to zero in $V_{\text{simulator}}$.
A smooth deformation field is encouraged by calculating
\begin{equation} \label{eq:5}
L_{\text{smooth}}=\frac{1}{N}\sum_{i=1}^{N}\frac{1}{\| \mathcal{N}(v_i) \|}\sum_{v_j\in\mathcal{N}(v_i)}\left \| u(i)-u(j) \right \|_{2},
\end{equation}
where $u(i)=c_{\text{simu}}(i)-c_{\text{norm}}(i)$ and $u(j)=c_{\text{simu}}(j)-c_{\text{norm}}(j)$ are respectively the displacement vectors of $v_{i}$ and $v_{j}$ on the simulated deformed bony surface. $\mathcal{N}(v_i)$ is the set of one-ring neighboring vertices of $v_{i}$. $N$ is the total number of vertices on the surface.
Finally, a $\ell_2$-norm regularization $L_{\text{simu\_reg}}$ of the network parameters is incorporated to avoid overfitting.

\subsubsection{Corrector Network:}
\label{sec:corrector_network}
\begin{figure}[!t]
	\centerline{\includegraphics[width=0.85\textwidth]{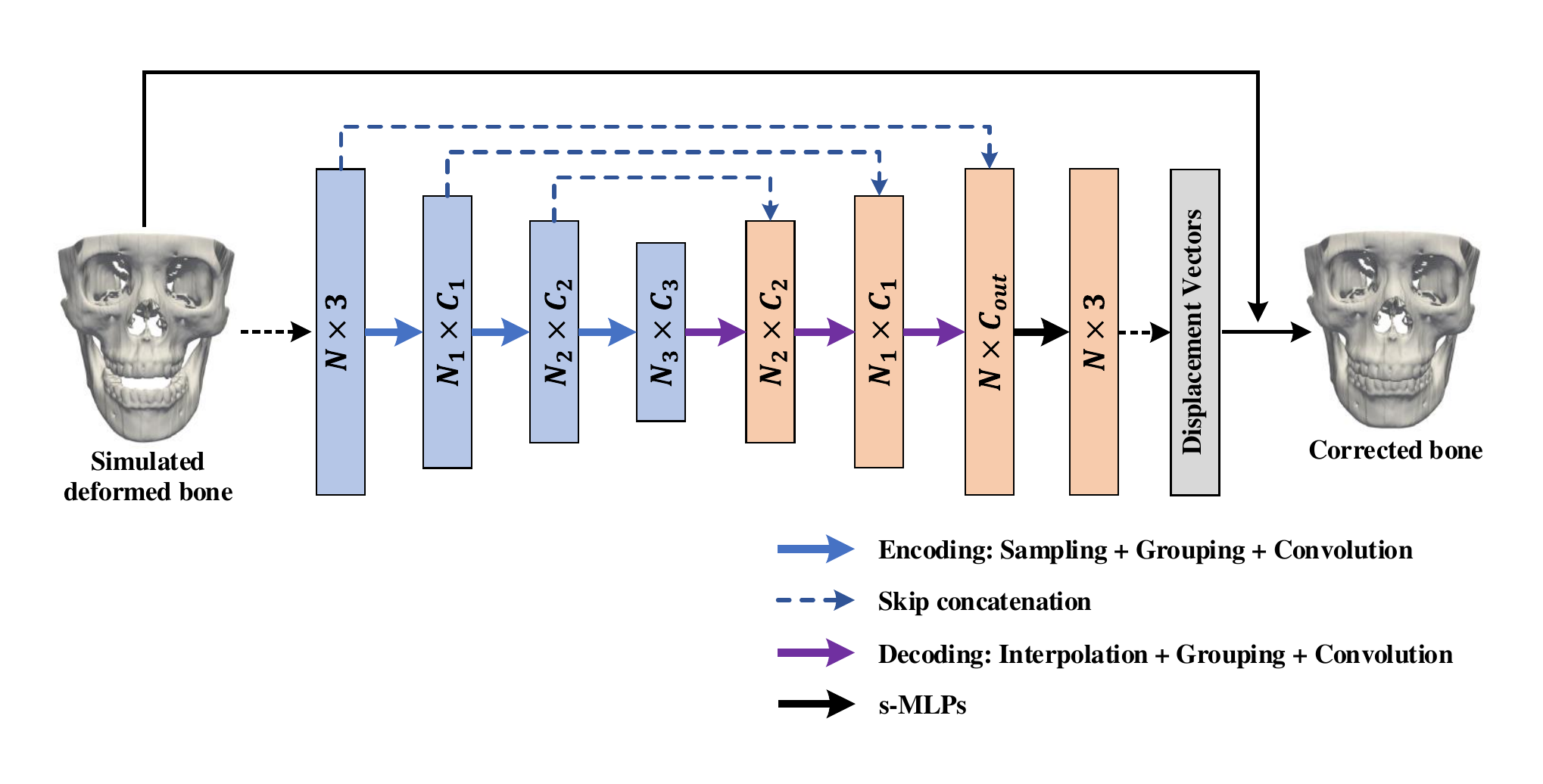}}
	\caption{The architecture of the corrector network with an encoding-decoding structure.} 
	\label{fig:corrector_network}
\end{figure}

The corrector is an encoding-decoding network (Fig.~\ref{fig:corrector_network}). Like the simulator network, each encoding layer in the corrector performs sampling, grouping, and convolution operations; each decoding layer performs interpolation, grouping, and convolution. The skip connection between two corresponding layers from the encoding and decoding streams is applied to learn point features integrating local and global shape information. Following the decoding stream, the point features are processed by a set of s-MLPs to obtain a $N \times 3$ vector, which corrects the positions of vertices of the simulated deformed bone to get a corrected bone.  

The loss function for the corrector network is
\begin{equation} \label{eq:6}
L_{\text{corrector}}=\frac{1}{N}\sum_{i=1}^{N}\left \| c_{\text{correct}}(i)-c_{\text{norm}}(i) \right \|_{2}+\lambda L_{\text{correct\_reg}},
\end{equation}
where $c_{\text{correct}}(i)$ and $c_{\text{norm}}(i)$ are the coordinate vectors of the $i$-th vertex on the corrected and input normal bony surface, respectively, $L_{\text{correct\_reg}}$ enforces $\ell_2$-norm regularization on the network parameters.

\subsubsection{Network Training and Inference:}
\label{sec:network_training_inference}
To train the proposed networks, vertex-wise correspondences of surfaces are established by non-rigidly matching a template surface with all bony surfaces in the training set. First, a group of corresponding landmarks localized on the training surfaces are rigidly aligned. From the aligned surfaces, we then choose the surface that is closest to the average landmarks as the template, and warp it to each rest surface via non-rigid coherent point drift (CPD) matching~\cite{Myronenko2010}. All warped surface templates are used for network training. We reduce the number of vertices via surface simplification~\cite{Garland1997} to reduce the computation cost. The vertex coordinates are min-max normalized. The loss functions are minimized with the Adam~\cite{Kingma2015} optimizer to determine the optimal network parameters. The two networks are alternatively trained in turn on each batch of samples. During the inference stage, the trained corrector is directly applied on a patient bony surface to estimate a reference bony shape model. The corrector network is invariant to the number and the order of vertices on the testing surface~\cite{Qi2017b}.

\section{Experimental Results}
\label{sec:experiments_results}
\subsubsection{Materials and Settings:}
\label{sec:materials_settings}
We used CT scans of 67 normal subjects from a previous study~\cite{Yan2010} and CT scans of 116 patients with jaw deformities to train our networks. The study was approved by our Institutional Review Board (\#Pro00009723). Following the clinical routine~\cite{Yuan2017}, all CT data were segmented to reconstruct 3D bony shape models. 51 landmarks were localized on each bony surface by experienced oral surgeons for calculating the loss function during the training only. Following Section~\ref{sec:network_training_inference}, a bony surface template from the training surfaces was selected based on the 51 landmarks to establish dense vertex correspondences. Ultimately, a total of 7772 (67 $\times$ 116) random pairs of deformed-normal bones were used to train the networks. 

The simulator network was implemented with 4 encoding, 4 fusion, and 4 decoding layers. The corrector network was configured with 4 layers each for encoding and decoding. The numbers of points for the encoding and decoding layers were $N_{\text{sub}}=\{\frac{N}{4},\frac{N}{16},\frac{N}{64},\frac{N}{128},\frac{N}{64},\frac{N}{16},\frac{N}{4},N\}$, where $N=4724$ during training, and $N>10\,000$ during testing. The radius $r$ of the 3D neighboring ball was set to $\{0.1,0.2,0.4,0.8\}$ and $\{0.8,0.4,0.2,0.1\}$ for the encoding and decoding layers, respectively. The output point features from the 4 encoding and 4 decoding layers had dimensions $C_{\text{feature}}=\{64,128,256,512\}$ and $C_{\text{feature}}=\{512,256,128,128\}$, respectively. Empirically, we set $\alpha=0.3$, $\beta=0.1$, and $\lambda=0.1$ in the loss functions \eqref{eq:1} and \eqref{eq:6}. The number $K=51$ of landmarks was set for calculating $L_{\text{jaw}}$ in \eqref{eq:4}. The networks were trained for 400 epochs with the learning rate of $0.0001$.

For testing, we acquired the data from another 24 patients with paired pre- and post-operative CT scans. The post-operative bones were used as ground truth for performance evaluation. Following the method in~\cite{Xiao2021}, we remeshed each post-operative bony surface according to its pre-operative bony surface to construct the vertex-wise correspondences for calculating estimation accuracy. Four evaluation metrics~\cite{Xiao2021}, i.e., vertex distance (VD), edge-length distance (ED), surface coverage (SC), and landmark distance (LD), were employed in the quantitative evaluation. We compared our method with DefNet~\cite{Xiao2021}, which was trained using 6834 paired samples simulated by sparse representation, 102 deformed bones were synthesized for each of 67 normal bones from 116 patient bones. Qualitative evaluation was performed by an experienced oral surgeon. Statistical significance of accuracy improvement was evaluated using the paired t-test. 

\subsubsection{Results:}
\label{sec:results}
\begin{figure}[t]
	\centerline{\includegraphics[width=1.0\textwidth]{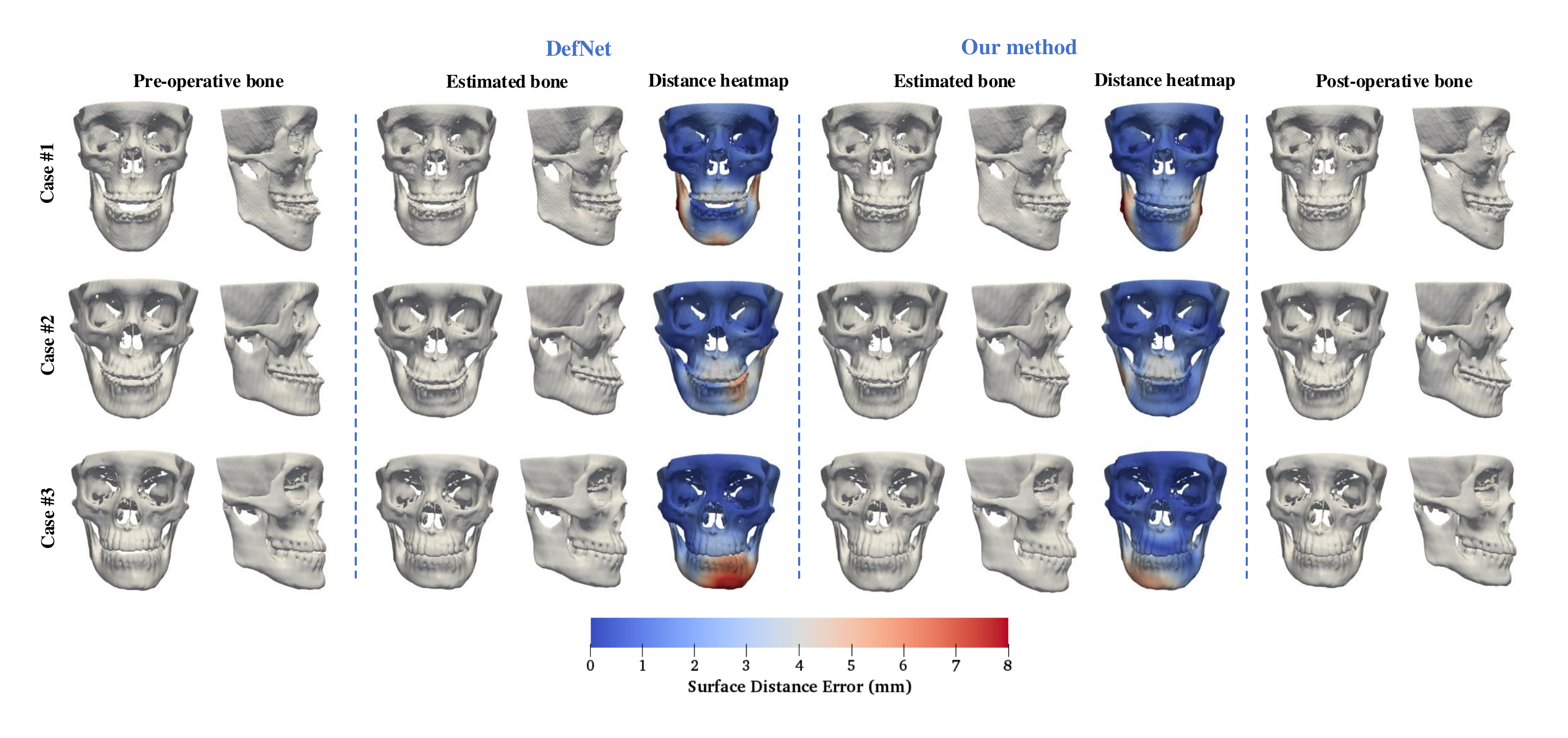}}
	\caption{Corrected bones estimated by the two methods for three randomly selected patients. Heatmaps of surface vertex distance are calculated by comparing the estimated bony surfaces with the corresponding post-operative bony surfaces (ground truth).} 
	\label{fig:qualitative_evaluation}
\end{figure}

The trained framework was tested on the data of 24 patients. Figure~\ref{fig:qualitative_evaluation} shows the estimated bony surfaces and the vertex-wise distance heatmaps for three randomly selected patients\footnote[1]{A video illustration of the three estimation examples is available as a supporting material.}. All results from our method inspected by the expert suggest that the estimated bones are clinically acceptable. For quantitative evaluation, the estimation accuracy for the jaw and the midface were calculated separately. Table~\ref{tab:quantitive_evaluation} shows that our method is statistically significantly more accurate in estimating the normal jaw ($p<0.05$) than DefNet based on the four metrics. The accuracy in maintaining the midface is comparable between the two methods ($p>0.05$). Overall, the proposed method outperforms DefNet in the term of accuracy.

Our method takes about 30 minutes for one training epoch, and 10 seconds for testing on each patient data, evaluated based on an NVIDIA Titan Xp GPU.

\begin{table}[!t]
	\centering
	\begin{threeparttable}
	\caption{Statistics for VD (mm), ED (mm), SC, and LD (mm) based on evaluation using 24 patients.}
	\setlength{\tabcolsep}{7pt}
	\renewcommand\arraystretch{1.1}
	\begin{tabular}{clcccc}
	\hline
	& \multicolumn{1}{c}{Method} & VD              & ED              & SC              & LD              \\ \hline
	\multirow{2}{*}{Jaw}     & DefNet                     & 4.08 $\pm$ 1.02 & 0.31 $\pm$ 0.06 & 0.69 $\pm$ 0.05 & 4.26 $\pm$ 0.95 \\
	& Ours                       & 3.49 $\pm$ 0.52 & 0.26 $\pm$ 0.05 & 0.72 $\pm$ 0.04 & 3.62 $\pm$ 0.53 \\
	\multirow{2}{*}{Midface} & DefNet                     & 1.36 $\pm$ 0.36 & 0.15 $\pm$ 0.04 & 0.96 $\pm$ 0.04 & 1.25 $\pm$ 0.31 \\
	& Ours                       & 1.39 $\pm$ 0.37 & 0.15 $\pm$ 0.04 & 0.95 $\pm$ 0.04 & 1.32 $\pm$ 0.36 \\ \hline
	\end{tabular}
	\begin{tablenotes}[flushleft]
		\small \item{Note: A larger SC indicates better estimation performance.}
	\end{tablenotes} 
	\label{tab:quantitive_evaluation}
	\end{threeparttable}
\end{table}

\section{Discussion and Conclusion}
\label{sec:discussion_conclusion}
As sufficient paired deformed-normal bones are almost impossible to acquire clinically, simulating paired data is necessary for training supervised deep learning models for our task. Data simulation based on sparse representation as done in DefNet \cite{Xiao2021} is limited in mimicking deformities, and requires significant efforts from experts to confirm simulation quality. Our self-supervised deep learning framework can be trained using unpaired data. The embedded simulator network has a stronger non-linear representation ability than sparse representation, and is able to accurately synthesize a broader range of jaw deformities based on patient bones. Compared with DefNet, our framework simplifies the training process and improves prediction accuracy.

\section*{Acknowledgement} This work was supported in part by United States National Institutes of Health (NIH) grants R01 DE022676, R01 DE027251, and R01 DE021863.

\end{document}